%% file: main.tex
\documentclass{article}
\usepackage{ijcai24}

\usepackage{times}
\usepackage{soul}
\usepackage{url}
\usepackage[hidelinks]{hyperref}
\usepackage[utf8]{inputenc}
\usepackage[small]{caption}
\usepackage{graphicx}
\usepackage{amsmath}
\usepackage{amsthm}
\usepackage{booktabs}
\usepackage{algorithm}
\usepackage{algorithmic}
\usepackage[switch]{lineno}
\usepackage{latexsym}
\usepackage{makecell}
\usepackage{float}
\usepackage{afterpage}

\usepackage[T1]{fontenc}
\usepackage{microtype}

\usepackage{inconsolata}

\usepackage{xcolor}
\definecolor{ocean}{RGB}{30,150,180}
\usepackage{multirow}

\usepackage{graphicx}
\graphicspath{ {./resources/} }

\title{GRASP: A Novel Benchmark for Evaluating Language GRounding and Situated Physics Understanding in Multimodal Language Models}

\author{
Serwan Jassim$^1$
\and
Mario Holubar$^2$\and
Annika Richter$^1$\and
Cornelius Wolff$^1$\and\\
Xenia Ohmer$^1$\And
Elia Bruni$^1$\\
\affiliations
$^1$Osnabrück University\\
$^2$University of Amsterdam\\
\emails
serwan.jassim@uos.de,
mario.holubar@student.uva.nl,\\
\{annrichter, cowolff, xenia.ohmer, elia.bruni\}@uos.de
}

\begin{document}
\maketitle

\input{sections/abstract}
\input{sections/introduction}
\input{sections/related_work}

\input{sections/methods}
\input{sections/results}

\input{sections/discussion}
\input{sections/conclusion}
\input{sections/acknowledgments}
\input{sections/contribution}

% Entries for the entire Anthology, followed by custom entries
\bibliographystyle{named}
\bibliography{ijcai24}

\end{document}

%% file: sections/abstract.tex
\begin{abstract}
    This paper presents GRASP, a novel benchmark to evaluate the language grounding and physical understanding capabilities of video-based multimodal large language models (LLMs). This evaluation is accomplished via a two-tier approach leveraging Unity simulations. The first level tests for language grounding by assessing a model's ability to relate simple textual descriptions with visual information. The second level evaluates the model's understanding of ``Intuitive Physics'' principles, such as object permanence and continuity. In addition to releasing the benchmark, we use it to evaluate several state-of-the-art multimodal LLMs. 
    Our evaluation reveals significant shortcomings in the language grounding and intuitive physics capabilities of these models. 
    Although they exhibit at least some grounding capabilities, particularly for colors and shapes, these capabilities depend heavily on the prompting strategy. At the same time, all models perform below or at the chance level of 50\% in the Intuitive Physics tests, while human subjects are on average 80\% correct.
    These identified limitations underline the importance of using benchmarks like GRASP to monitor the progress of future models in developing these competencies.
\end{abstract}

%% file: sections/introduction.tex
\section{Introduction}\label{sec:introduction}

The remarkable progress of large language models (LLMs) has sparked intense debate about their potential for achieving genuine machine cognition and human-level comprehension. A very important question is whether these models possess an understanding of the physical world, even though they learn primarily through next-word prediction, so in a fundamentally ungrounded manner. Recently, multimodal LLMs have started to make their mark, learning not just from vast text databases but also from extensive visual inputs. However, it remains unclear whether multimodality enhances language grounding and physical understanding. With \textbf{GRASP}, which stands for \textbf{GR}ounding \textbf{A}nd \textbf{S}ituated \textbf{P}hysics, we propose a novel evaluation benchmark to answer these questions for video-based LLMs.

Most research on physical understanding in LLMs has been centered around tasks that assess a model's ability to correctly link a textual question to its corresponding textual answer~\cite{gordon-etal-2012-semeval,hendrycks2021measuring}. For instance, consider a data point in the “Choice of plausible alternatives” (COPA) dataset~\cite{gordon-etal-2012-semeval}: Given the text “I poured water into the glass.”, the model has to choose one of the two effects “The water quenched my thirst.” or “The glass became full.” Under a conservative interpretation, this test evaluates whether the model has learned to map the input text pattern onto the correct output text pattern from examples of similar sentence pairs in its vast training dataset. It certainly does not evaluate whether the model knows what water looks and behaves like and whether the model genuinely understands that pouring liquid into a glass will fill up that glass. 

GRASP advances the evaluation of language grounding and physical grounding by leveraging multimodality: Questions about objects and their behaviors are connected to an external environment. 
Specifically, we use the Unity\footnote{\url{https://unity.com/}} game engine to simulate various scenes (in the form of videos) as a proxy for the real world. 
The model has to answer textual questions about these scenes, which are designed to assess its ability to relate text to the physical world (\emph{grounding}) as well as its ability to understand the basic principles of physics that govern how objects behave in the world (\emph{Intuitive Physics}). 

Based on these two aspects, the GRASP dataset is comprised of two levels. 
Level 1 tests for grounding: The model has to detect simple objects and recognize object features, relative positions, and (direction of) motion. This level tests whether a model can map between simple textual descriptions and visual inputs. 
Passing Level 1 is necessary, if the model is to generate statements about the physical properties and behaviors of objects in a visual scene. 
It is therefore a requirement for passing Level 2 which tests for Intuitive Physics: The model has to judge the physical plausibility of video sequences, where simple objects behave according to, or violate, Intuitive Physics concepts such as \textit{object permanence} or \textit{continuity}.
The knowledge and abilities of LLMs are tied to their accessibility through language. 
Thus, the two levels evaluate a model's ability to ``perceive'' the environment and to ``reason'' about the physical events therein within the constraints of its language interface.

This language interface differentiates GRASP from other image- or video-based datasets for Intuitive Physics, which were largely developed to train and evaluate dedicated Intuitive Physics models (lacking language) ~\cite{battaglia_interaction_networks,watters_visual_interaction_networks,Piloto2022}. 
Most relevant to our approach is the recently developed Physical Concepts\footnote{\url{https://github.com/deepmind/physical_concepts}} dataset~\cite{Piloto2022}, which, similar to our Level 2, consists of simulated videos of physically plausible and implausible events.  
Instead of targeting dedicated Intuitive Physics models, GRASP is designed to evaluate whether language grounding and Intuitive Physics are a subset of the abilities that emerge in multimodal LLMs.

To this end, GRASP extends and improves existing datasets in several ways. First, we introduce novel stimuli to test for grounding (Level 1). While grounding is naturally absent in vision-only models it is a key prerequisite for a (question-based) evaluation of Intuitive Physics in multimodal LLMs. 
Second, we significantly broaden the range of Intuitive Physics concepts that can be tested (Level 2).
While \textit{Physical Concepts} comprises five concepts, our benchmark comprises eight and provides multiple experiments for some concepts as well as combinations of concepts. 
Lastly, by disseminating the Unity source code along with the benchmark data, we enable the community to customize and expand the benchmark as multimodal LLMs become more sophisticated.\footnote{The benchmark including all our code as well as supplementary material is available at \url{https://github.com/i-machine-think/grasp}.}

Accompanying the release of GRASP, we provide scores for five state-of-the-art multimodal LLMs.
Our findings reveal that, despite their impressive capabilities, current multimodal LLMs are still lacking in both language grounding and intuitive physics understanding. 
While the tested models demonstrate certain grounding capabilities, specifically regarding colors and shapes, they universally fail the Intuitive Physics tests.
These shortcomings emphasize the necessity for using benchmarks like GRASP to monitor the progress of future models in terms of these capabilities.

%% file: sections/related_work.tex
\section{Related Work}\label{sec:related_work}

GRASP takes inspiration from psychology research on \textit{Intuitive Physics in early development}. It is related to other datasets that have been created to develop \textit{neural network models of Intuitive Physics} but targets language models. We study \textit{grounding and physics understanding in LLMs} but instead of looking at text-based models, we focus on recently developed \textit{video-based multimodal LLMs}.

\paragraph{Multimodal LLMs.}
%\xo{If we need more space, we can remove this first paragraph in my opinion. It is very general and not directly related to our work.}
%In recent years, LLMs have demonstrated remarkable capabilities in language understanding and reasoning tasks.
%Following the success of early foundation models like GPT-2 \cite{radford2019language} and BERT \cite{devlin2019bert}, development has moved towards a steady increase in model size and training data.
%Models with hundreds of billions of parameters, such as GPT-3 \cite{brown2020language}, PaLM \cite{chowdhery2022palm}, Gopher \cite{rae2022scaling} and BLOOM \cite{workshop2023bloom} have exhibited exceptional generalization and in-context learning performance \cite{brown2020language,wei2022emergent}.

With the recent widespread success of LLMs, researchers have also explored their use for processing multi-modal inputs. A key idea utilized by seminal works such as Flamingo \cite{alayrac2022flamingo} and BLIP-2 \cite{li2023blip2} is to align a pretrained vision model with the textual embedding space of an LLM. LLaVA \cite{liu2023visual} and MiniGPT-4 \cite{zhu2023minigpt4} combine this technique with instruction tuning to deliver an end-to-end chatbot with visual reasoning abilities.
 Recent work has extended this approach to video data. VideoChat \cite{li2023videochat} was among the first to do so, followed by Video-ChatGPT \cite{maaz2023videochatgpt}. Video-LLaMA \cite{zhang2023videollama} adds an audio processing branch to the pipeline, and PandaGPT \cite{su2023pandagpt} supports a total of six different modalities including thermal images and IMU sensor data.
Although these models differ in implementation and training data, they are all based on the premise of aligning the embeddings of pretrained foundation models through a learnable interface.

\paragraph{Intuitive Physics in early development.}
To evaluate Intuitive Physics in LLMs, we rely on extensive research on the subject from developmental psychology. 
In particular, we focus on fundamental Intuitive Physics concepts that have been identified in this research, such as object permanence~\cite{baillargeon_permanence,baillargeon_1987,spelke_1992},
%occlusion~\cite{Hespos2001-bi},
gravity~\cite{kim_1992,kim_1999,spelke_1992}, and inertia~\cite{kim_1999,spelke_1992,spelke_1994}.
Studies in this field typically employ simple experimental setups with geometric shapes that are part of physically possible or physically impossible events and assess understanding with the so-called \textit{violation-of-expectation} (VoE) paradigm.
This paradigm is based on the idea that infants will show surprise---measured through their behavioral or physiological response---when an event violates their expectations. 
We draw inspiration from known experimental setups to develop the videos for GRASP. 
Since LLMs possess language, we can inquire about the plausibility or implausibility of the scenes directly, mapping the VoE paradigm onto a binary classification problem.

\paragraph{Neural network models of Intuitive Physics.}
Advances in deep learning and AI have paved the way for the development of dedicated models capable of learning Intuitive Physics from data (for an overview, see Duan et al.~\shortcite{ijcai2022p0763}).
One of the earliest examples is the so-called \textit{Interaction Network}~\cite{battaglia_interaction_networks}. It received explicit information about objects and their relationships for a given scene and was trained on next-state prediction.
Scenes included n-body systems, bouncing balls, and springs colliding with rigid bodies. 
Since then, the research focus has shifted to learning Intuitive Physics from raw visual inputs. 
For example, Watters et al.~\shortcite{watters_visual_interaction_networks} extended the Interaction Network with a convolutional neural network to process image inputs.
Other prominent examples include neural networks predicting the behavior of block towers~\cite{pmlr-v48-lerer16} or the dynamics of robot-object interactions~\cite{agrawal_NIPS2016}.

As a part of model development and evaluation, several Intuitive Physics benchmarks have been developed. 
Among others, the Physical Concepts~\cite{Piloto2022}, IntPhys~\cite{intphys}, InfLevel~\cite{weihs2022benchmarking}, and AVoE~\cite{dasgupta2021benchmark,dasgupta2021aVoE} benchmarks utilize simulated videos to assess physical concepts in computer vision models. 
Like GRASP, they focus on established concepts and VoE experiments from the developmental psychology literature.
GRASP can be considered an expansion of these benchmarks: It adds an entirely new level to test for grounding, as well as more concepts, more scenes per concept, and combinations of concepts. 
Besides, unlike all the approaches above, it is designed to evaluate LLMs and therefore uses a language interface, allowing for analyses within and across the modalities of language and vision.

\paragraph{Grounding and physics understanding in LLMs.}
Before the advent of multimodal LLMs, it has often been argued that LLMs---being trained on text---fail to relate language to the physical world~\cite{bisk-etal-2020-experience}. Progress on neural networks with grounded language abilities largely happened in research fields originating from image captioning~ \cite{mitchell-etal-2012-midge}, such as visual question answering (VQA)~\cite{Antol_2015_ICCV}, instruction following~\cite{Anderson_2018_CVPR,Das_2018_CVPR}, and visual commonsense reasoning~\cite{Zellers_2019_CVPR}. 
Among others, the VQA datasets CLEVR~\cite{clevr} and CLEVRER~\cite{Yi2020CLEVRER} have been developed in the context of this research. The former tests for compositional language and elementary visual reasoning using images of 3D shapes. The latter tests for various visual reasoning capabilities based on videos of colliding objects (again 3D shapes).
At the time CLEVRER was released, SOTA models performed well on descriptive tasks but poorly on causal ones requiring explanation, prediction, or counterfactual reasoning. 
Now, multimodal LLMs promise to provide extensive language capabilities while being able to relate between text and images or videos. For example, GPT-4 outperforms SOTA models on various VQA benchmarks (\url{https://openai.com/research/gpt-4}). 
GRASP provides a novel, and extensive benchmark to test fundamental grounding and Intuitive Physics skills in a question-answering setting.

%% file: sections/methods.tex
\section{Benchmark Design}\label{sec:methods}
GRASP is a two-level benchmark, with each level containing multiple visual tests.
These tests were modeled in the Unity simulator and compiled into a dataset in the form of videos.\footnote{Although most Level 1 tests do not entail dynamics and could therefore be represented as images, we decided to capture them as videos to ensure consistency across all Level 1 and 2 tests.}
All videos are ten seconds long and were generated at 50 frames per second.

\subsection{Level 1 (Grounding)}\label{subsec:level1dataset}
\begin{figure}
\centering
\includegraphics[width=0.95\columnwidth]{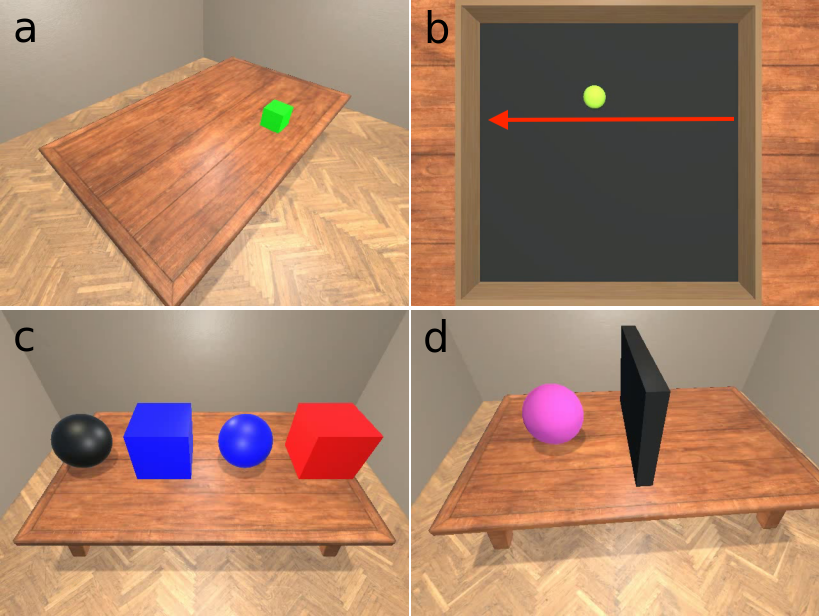}
\caption{Examples from GRASP's Level 1: (a) Shape \& Color (both have the same setup, but differ in the randomized property), (b) Directionality \& Movement (ball is rolling from right to left as indicated by red arrow), (c)~Object Ordering, (d) Relational Position.}
\label{fig:level1}
\end{figure}

The initial stage of GRASP evaluates the elementary visual understanding capabilities of LLMs. This stage comprises tests that assess basic visual comprehension and lay the groundwork for higher-order reasoning required in the subsequent level. We premise that models struggling at this foundational phase will likely encounter difficulties in the next stage, where they must discern and reason about more complex physical interactions. This approach ensures a sequential increase in task complexity, aligning with the natural progression of cognitive development. More specifically, Level 1 comprises six test categories:
\begin{itemize}
    \item \textbf{Shape}: A cube or a sphere of random size is spawned at a random location on a table.
    \item \textbf{Color}: A black, blue, green, or red sphere of random size is spawned at a random location on a table.
    \item \textbf{Directionality}: A ball rolls forward, backward, right, or left.
    \item \textbf{Movement}: A ball rolls in a random direction or stands still.
    \item \textbf{Object Ordering}: A random sequence of balls or cubes (between two and four) of random color are spawned on a table. Each object in one video is unique.
    \item \textbf{Relational Position}: A ball is spawned either to the left or right of a barrier.
\end{itemize}
For each test, we generate 128 videos.
Examples of these tests are displayed in Figure~\ref{fig:level1}.
Importantly, the elements used in this phase---such as objects in various shapes, colors, and positional relationships---form the fundamental components of the videos in the next stage.

\subsubsection{Prompts}\label{subsec:level1prompts}
For this level of the benchmark, we introduce two different sets of prompts, each giving rise to a different classification task.
To maintain uniformity with Level 2, one set is designed to induce a \textbf{binary classification} problem. 
We generate positive and negative samples by combining each input video with a prompt that proposes an observation and asks whether this observation is true.
The observation is either a true (pos. sample) or a false (neg. sample) statement about the video.
In Figure~\ref{fig:level1}c, for example, the model is prompted with ``From left to right, the following objects are on the table: black ball, blue cube, blue ball, red cube. Is this true?'' (pos. sample) and ``From left to right, the following objects are on the table: red cube, blue ball, black ball, blue cube. Is this true?'' (neg. sample).
A complete list of the prompts is provided in the supplementary material.
For the Object Ordering test, we exclusively create negative samples by permuting the existing objects, to specifically assess ordering accuracy. For all other tests, we sample entirely different observations.

\begin{figure*}[ht]
\centering
\includegraphics[width=\textwidth]{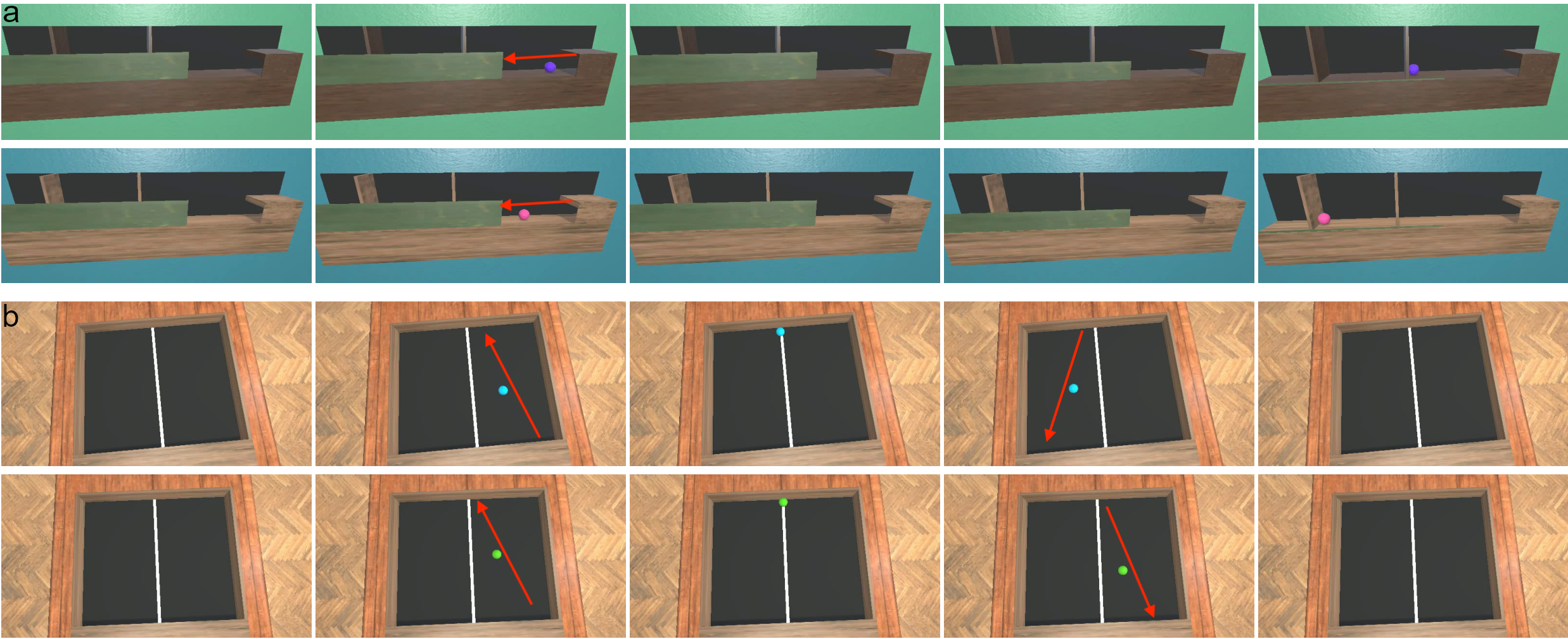}
\caption{Examples from GRASP's Level 2. The first and second rows display plausible and implausible versions of each experiment, respectively. The examples here test the understanding of Solidity \& Continuity (a), and Inertia (b). We cut off some of the background here.}
\label{fig:level2}
\end{figure*}

To evaluate the models' sensitivity to the instructions, we propose a second set of prompts that allow for open-ended answers and frame the problem as \textbf{multi-class classification}.
For example, the prompt for the Directionality test is ``Which direction is the ball rolling?'' in this case.
These prompts were not introduced for Object Ordering since it proved too difficult to parse the answers and also not for Movement which does not allow for an open-ended question.
All open-ended question prompts can be found in the supplementary material.

\subsection{Level 2 (Intuitive Physics)}\label{subsec:level2dataset}
Level 2 comprises tests of Intuitive Physics understanding, which follow the structure of VoE experiments by contrasting physically plausible and implausible scenes.
Specifically, Level 2 comprises tests for the following eight concepts and events:
\begin{itemize}
    \item \textbf{Continuity}: Objects cannot teleport in space and time, they can only move along continuous paths~\cite{spelke_1992}.
    \item \textbf{Solidity}: Objects cannot overlap in space and time, they can only move along clear paths~\cite{spelke_1992}.
    \item \textbf{Unchangeableness}: Objects cannot spontaneously change their size, shape, and color~\cite{baillargeon_2012_2}.
    \item \textbf{Gravity}: Objects move downward without existing support~\cite{spelke_1992}.
    \item \textbf{Support}: Objects maintain stability when on a platform, but lose stability when positioned off it~\cite{baillargeon_1995}.
    \item \textbf{Collision}: Objects get displaced when hit by moving objects~\cite{baillargeon_1995}.
    \item \textbf{Object Permanence}: Objects continue to exist when they are occluded~\cite{baillargeon_1995}.
    \item \textbf{Inertia}: Objects do not spontaneously alter their motion~\cite{spelke_1992}.
\end{itemize}
We adopt 16 tests from the psychology literature and create a dataset containing 128 physically plausible and 128 physically implausible videos for each of them.
%Figure~\ref{fig:level2} provides illustrations for two of these tests (see Appendix~\ref{sec:level2_tests}, for the other tests) and Table~\ref{tab:level2_distribution} indicates how the tests are distributed across the different physical concepts and events.
Illustrations of these tests can be found in Figure~\ref{fig:level2} and the supplemental material and their distribution across the physical concepts and events is provided in Table~\ref{tab:level2_distribution}.
Since it is not always possible to strictly disentangle all the concepts in the tests (e.g. in the implausible scenario displayed in Figure~\ref{fig:level2}a, the ball could have either teleported behind (continuity violation) or rolled through the barrier (solidity violation)), there is a discrepancy between the sum of the distribution in Table~\ref{tab:level2_distribution} and the number of tests.
We also adopt the use of occluders to hide physical manipulations from experimental studies (e.g. Figure~\ref{fig:level2}a).
This ensures that models need to infer plausibility using the entire history of a video instead of individual frames.

\subsubsection{Visual Randomization}\label{subsec:visualrandomization}
In GRASP, models are validated against multiple videos per test to ensure that results are representative.
We achieved rich visual variation for each test by randomly sampling object colors and textures, background colors, camera angles, movement speeds, start delays, orientations of experimental setups, as well as other test-specific parameters (see supplementary material) for each video, as applicable.

\begin{table}[t]
    \begin{center}
    \begin{tabular}{cccc}
        \toprule
        Continuity & Solidity & Inertia & Gravity \\
        \midrule
        4 & 2 & 4 & 5 \\
        \bottomrule
        \toprule
        Collision & OP & Support & Unchangeableness \\
        \midrule
        1 & 3 & 1 & 2 \\
        \bottomrule
    \end{tabular}
    \end{center}
    \caption{Test distribution over physical principles/events that are evaluated in Level 2 of GRASP (OP = Object Permanence).}
    \label{tab:level2_distribution}
\end{table}

\subsubsection{Prompts}\label{subsec:level2prompts}
The model has to evaluate the physical plausibility of the videos, that is, perform a binary classification task.
In contrast to Level 1, negative and positive samples of a test share the same prompt and are distinguished by the video contents.
All prompts are based on the same template:
\begin{quote}
        The video you're seeing was generated by a simulator. Given how objects behave on earth, is \mbox{$\langle$\textit{observation}$\rangle$} plausible? Your answer should be based on the events in the video and ignore the quality of the simulation. Answer only with yes or no.
\end{quote}
In this template, $\langle$\textit{observation}$\rangle$ denotes a phrase that hints at what the model should pay attention to when inferring physical plausibility, e.g. ``the trajectory of the ball'' or ``the final position of the ball''.
The models are instructed to ignore details related to the simulation quality to prevent their judgment from being influenced by visual inaccuracies.
A complete list of prompts can be found in the supplementary material.

\section{Experiments}\label{sec:experiments}

In our experiments, we evaluate several state-of-the-art models as well as human subjects on our benchmark. Details on the model evaluation are provided in Sections \ref{subsec:models}--\ref{subsec:exp_setup} and details on the human evaluation in Section \ref{subsec:human_trial}.

\subsection{Models}\label{subsec:models}

We consider several multimodal LLMs that can perform video-based question answering.

\textbf{Video-ChatGPT} \cite{maaz2023videochatgpt} leverages LLaVA \cite{liu2023visual} as a vision-language model and adapts it to video data, fine-tuning the model on a dataset of video-instruction pairs to enable understanding of temporal dynamics. Specifically, it uses LLaVA-Lightning-7B v1.1, which is comprised of CLIP ViT-L/14~\cite{radford2021learning}, as a visual encoder and Vicuna-7B v1.1 \cite{vicuna2023} as a language decoder.

\textbf{Video-LLaMA} \cite{zhang2023videollama} enables simultaneous visual and auditory understanding using a multi-branch cross-modal pre-training framework. The vision-language branch uses CLIP ViT-G/14~\cite{radford2021learning} and BLIP-2 Q-Former \cite{li2023blip2} and is trained using video-text as well as image-text data. We test three versions of the model, which use Vicuna-7B v0, Vicuna-13B v0, and LLaMA-2-7B as their respective language decoder. They are all fine-tuned on instruction-tuning data from MiniGPT-4, LLaVA, and VideoChat.

\textbf{PandaGPT} \cite{su2023pandagpt} uses the joint embeddings of ImageBind \cite{girdhar2023imagebind} to enable a Vicuna model to reason about image, video, depth, thermal, and IMU data. The multimodal encoder's feature space is aligned with the language model by training on image-language instruction-following data. The particular versions we test are the 7B version with a maximum sequence length of 1024 using Vicuna-7B v0 and the 13B version with a maximum sequence length of 400 using Vicuna-13B v0.

\textbf{VTimeLLM} \cite{huang2023vtimellm} adds an additional stage to the training pipeline alongside feature alignment and instruction tuning. This stage, called Boundary Perception, aims to improve the model's temporal understanding abilities by training on a dataset of time-segmented and event-annotated videos. It uses CLIP ViT-L/14 as its visual encoder and Vicuna-7B v1.5 as the language decoder.

\subsection{Prompting}\label{subsec:prompting}
Apart from the prompts introduced in Section~\ref{sec:methods}, we report the results for additional prompting strategies for Level 1. 
For Level 2 changes to the prompting strategy did not impact the results (the models still fail to perform the task).
In particular, we include one-shot prompting and chain-of-thought (CoT) prompting for the Level 1 binary classification task.
%In addition to the prompts introduced in Section~\ref{sec:methods}, we also employ one-shot and chain-of-thought (CoT) prompting for the Level 1 binary classification task.
%\footnote{One-shot and CoT prompting is omitted from Level 2 since it did not have any impact on the results.}
With one-shot prompting, we ``familiarize'' the models with the task by prepending an example question-answer pair to the main question.
For instance, a one-shot prompt for the Color test is:
``This is an example of a question about this video and the correct answer.
Question: The ball on the table is red. Is this true? Answer only with yes or no. Answer: Yes. Next, I want you to answer my next question in the same way with regard to the next video.'' 
Whether a positive or negative sample is used in the initial prompt is randomized.
For CoT prompting, we prepend the open-ended question (see supplementary material) to the binary classification prompt (for the Movement and Ordering tests, we use the question ``What can you see in this video?''). For both, CoT and one-shot prompting, we allow the model to reply before submitting the final instruction containing the task.

\subsection{Experimental Setup}\label{subsec:exp_setup}
Experiments are conducted on an Nvidia 3090 GPU for the 7B models and on an Nvidia A100 for the 13B models.
For all models, we use their default parameters but adapt the system prompt when applicable (see supplementary material for details).
Each video-prompt pair is evaluated three times with a different seed per model.
For quantitative evaluation, the models' responses are classified by a simple scheme:
Responses to binary yes/no questions are only counted as valid if they begin with the word ``yes'' or ``no''; the rest of the response is considered irrelevant.
We regard responses that do not adhere to this as incorrect.
For open-ended questions in Level 1, we use a parsing scheme that also considers slight deviations from the ground truth as correct (e.g. ball and sphere are considered to be equivalent).
The full parsing scheme is outlined in the supplementary material.

\begin{table*}[t]
    \begin{center}
    \scalebox{0.8}{
    \begin{tabular}{cc||c|c|c|c|c|c|c}
        \toprule
        \textbf{Task} &  \textbf{Test} & 
        \parbox{2cm}{\centering\small\textbf{Video-LLaMA \\(7B)}} & 
        \parbox{2cm}{\centering\small\textbf{Video-LLaMA \\(13B)}} & 
        \parbox{2.2cm}{\centering\small\textbf{Video-LLaMA2 \\ (7B)}} &  
        \parbox{1.5cm}{\centering\small\textbf{PandaGPT \\ (7B)}} & 
        \parbox{1.5cm}{\centering\small\textbf{PandaGPT \\ (13B)}} & 
        \parbox{1.7cm}{\centering\small\textbf{VTimeLLM \\ (7B)}} & 
        \parbox{2.3cm}{\centering\small\textbf{Video-ChatGPT \\ (7B)}} \\
        \midrule
        \midrule
        \multirow{6}{*}{\makecell{Binary \\ Classification \\ Zero-Shot}} & Shape &  49.1 &  49.1 & 48.3 & 50.0 &  50.0 & 50.0 & 49.9 \\
        & Color &  49.6 &  52.1 & 50.8 & 50.0 &  50.0 & 50.0 & 50.1 \\
        & Movement &  49.6 &  46.6 & 49.6 & 35.2 &  50.0 & 50.0 & 50.4 \\
        & Directionality &  48.7 & 45.8 &  49.9 & 47.3 &  50.0 & 50.0 & 49.6 \\
        & Relational Position &  50.1 &  47.9 & 49.0 & 50.0 &  50.0 & 50.0 & 49.7 \\
        & Ordering (avg.) &  49.6 & 49.7 &  50.0 & 50.0 &  50.0 & 50.0 & 50.3 \\
        \midrule
        \multirow{6}{*}{\makecell{Binary \\ Classification \\ CoT}} & Shape &  \textbf{69.4} & \textbf{ 69.5} & \textbf{63.8} & 33.6 &  \textbf{80.9} & 50.0 & \textbf{70.1} \\
        & Color &  \textbf{79.7} &  \textbf{73.3} & \textbf{84.2} & \textbf{76.2} &  \textbf{70.7} & 50.0 & \textbf{65.2} \\
        & Movement &  48.8 &  48.8 & 48.3 & 44.1 &  35.9 & 50.0 & 43.1 \\
        & Directionality &  45.3 & 47.5 &  50.1 & 51.2 &  51.2 & 50.0 & 46.1 \\
        & Relational Position &  46.9 &  48.7 & 51.4 & 50.0 &  50.0 & 50.0 & 47.4 \\
        & Ordering (avg.) &  48.6 & 49.0 &  50.0 & 49.9 &  50.7 & 50.0 & 50.6 \\
        \midrule
        \multirow{6}{*}{\makecell{Binary \\ Classification \\ One-Shot}} & Shape &  41.4 &  23.6 & 45.6 & 50.5 &  46.1 & 41.9 & 48.4 \\
        & Color &  43.6 &  23.0 & 49.6 & 29.6 &  44.3 & 32.0 & 49.0 \\
        & Movement & 37.1  &  26.3 & 38.0 & 28.0 &  39.1 & 37.5 & 50.8 \\
        & Directionality &  36.6 & 20.1 &  42.8 & 37.4 &  42.8 & 33.5 & 50.7 \\
        & Relational Position &  40.2 &  21.0 & 37.4 & 53.4 &  35.4 & 25.7 & 51.4 \\
        & Ordering (avg.) &  42.2 & 22.3 &  22.2 & 50.7 &  50.5 & 20.5 & 49.7 \\
        \midrule
        \midrule
        \multirow{4}{*}{\makecell{Multi-Class\\Classification\\Zero-Shot}} & Shape &  \textbf{87.2} &  \textbf{83.3} & \textbf{75.8} & 49.2 &  14.8 & 14.1 & 14.3 \\
        & Color &  \textbf{93.2} &  \textbf{74.2} & \textbf{90.9} & \textbf{70.3} &  \textbf{73.4} & \textbf{85.7} & \textbf{76.0} \\
        & Directionality &  14.1 & 15.9 &  11.2 & 26.6 &  25.0 & 20.8 & 17.4 \\
        & Relational Position &  28.9 &  24.2 & 35.4 & 0.0 &  50.0 & 49.7 & 49.5 \\
        \bottomrule
    \end{tabular}
    }
    \end{center}
    \caption{Accuracy (\%) for all models on GRASP's Level 1 using binary question  (inducing binary classification) and open-ended question prompts (inducing multi-class classification). For binary classification, results are listed for zero-shot, one-shot, and chain-of-thought (CoT) prompting strategies.  We highlight all accuracies that lie notably above chance performance. For binary classification, an accuracy of 50\% coincides with chance performance, while for multi-class classification it is 50\% for Shape and Relational Position, and 25\% for Color and Directionality.}
    \label{tab:level1_results}
\end{table*}

\subsection{Comparison with Human Subjects}\label{subsec:human_trial}
To validate and assess the difficulty of our benchmark, we submit GRASP's Level 2 tests to AWS Mechanical Turk\footnote{\url{https://www.mturk.com}} for a human trial.
We focus our evaluation on Level 2, considering that human subjects can trivially solve tests in Level 1. 
In our experiment, participants are asked to judge the physical plausibility of each Level 2 test, resulting in 16 videos per questionnaire.
For each test, we randomly sample whether to serve the plausible or implausible scene.
Furthermore, we randomize the order in which the videos are displayed to each participant.
Three independent submissions are collected per video from which a final answer is determined using a majority vote.
We collect submissions from 120 participants, i.e. a sub-set of 40 videos are being classified per Level 2 test (20 per plausible and 20 per implausible scene).

%% file: sections/results.tex
\section{Results}\label{sec:results}
Table~\ref{tab:level1_results} presents accuracy scores for the multimodal LLMs tested on \textbf{Level 1 (grounding)}.
For zero-shot binary classification, the average scores across all models and tests indicate a performance close to chance (50\%).
Results below chance performance can be attributed to ambiguous answers that could not be parsed.
Examining scores on positive and negative samples individually (see supplementary material) highlights distinctive behaviors among the models.
Video-LLaMA exhibits a consistent bias toward responding ``yes'' to prompts, while Video-ChatGPT displays a more dynamic bias, shifting between ``yes'' and ``no'' responses across different tests.
PandaGPT and VTimeLLM, in turn, consistently respond with ``yes'' regardless of the test category.

Furthermore, the results show that binary classification with CoT prompting leads to a consistent improvement above chance performance for the Shape and Color tasks across all models, except for PandaGPT (7B) and VTimeLLM.
On the other hand, such an improvement is not observed when running binary classification with one-shot prompting.
In this case, models tend to perform even worse than with zero-shot prompting due to simply repeating the answer from the provided example.

Accuracies for multi-class classification in Level 1 are also presented in Table~\ref{tab:level1_results}.
For the Directionality and the Relational Position tasks, the performance coincides with zero-shot binary classification for all models, being either chance or below chance due to ambiguous answers (in this case, accuracies of 25\% and 50\% equal chance performance for Directionality and Relational Position respectively).
Similar to the CoT results, all models perform quite well in the Color task as well as Video-LLaMA and PandaGPT (7B) in the Shape task.

\begin{figure}[ht]
\includegraphics[width=\linewidth]{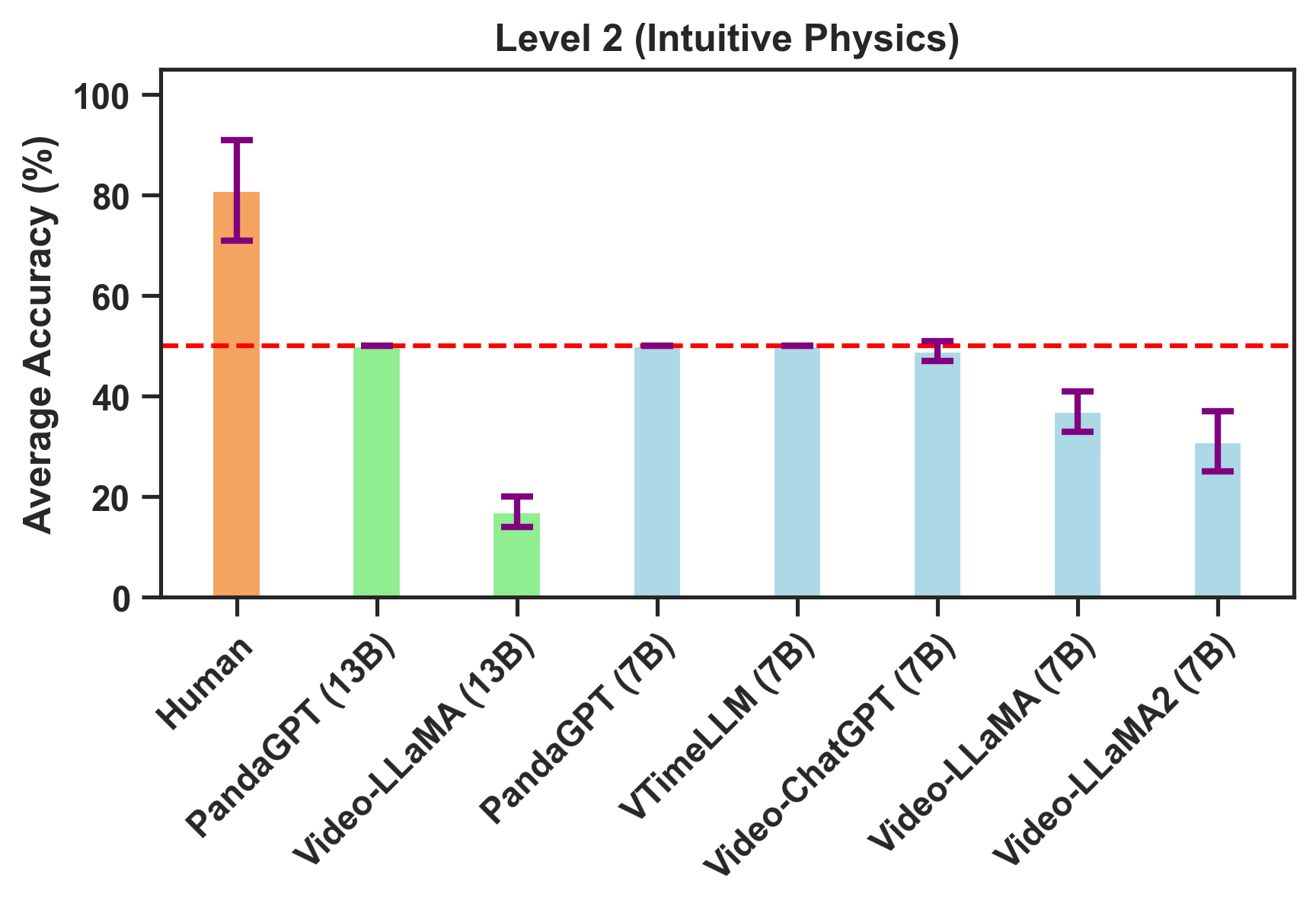}
\caption{Average accuracies (\%) over Level 2 tests for all models and the human trial. The red line indicates chance performance and error bars represent the standard deviation over all tests (positive and negative scenes combined). Models with green bars have 13B parameters and models with blue bars have 7B parameters.}
\label{fig:level2_results}
\end{figure}

Results for \textbf{Level 2 (Intuitive Physics)} are displayed in Figure~\ref{fig:level2_results}.
We do not report individual scores since the models generally exhibit performance equivalent to, or less than, chance across all tests as indicated by the error bars.
Video-LLaMA performs significantly below chance because it generates answers that cannot be parsed.
The models' poor performance on Level 2 is not surprising given that they already failed on Level 1, which assesses basic grounding abilities necessary to answer questions about the videos at Level 2.
With human performance at approximately 80\%, the results suggest that while the task is solvable, it presents a non-trivial challenge.
In particular, difficulties in the human trial were observed in one test concerning the inertia principle where participants were not able to correctly observe a discrepancy between the incoming and outgoing angle of a deflection.
This test will be highlighted accordingly in the published benchmark such that researchers can decide whether to include it in their evaluations.
%In particular, difficulties in the human trial were observed in the implausible Inertia, Gravity \& Inertia, and Object Permanence tests (see Tests 6, 9, and 10 in Appendix~\ref{sec:level2_tests} respectively), with an accuracy of 35\%, 50\%, and 50\% respectively.

To understand whether the bad performance on Level 2 is due to the multimodal nature of the task, we additionally evaluate Video-LLaMA on textual descriptions of the test events (see supplementary material). Video-LLaMA and Video-LLaMA2 are the only models out of those tested that allow for text-only inputs. 
On the binary classification task, the 7B and 13B parameter versions of Video-LLaMA achieve an accuracy of 49.8\% and 53.7\%, respectively, while Video-LLaMA2 achieves 51.9\%.
In comparison, GPT-4 achieves an accuracy of 75.0\% on this task.
This analysis suggests that video-based LLMs might also lack an Intuitive Physics understanding when evaluated on text alone.

%% file: sections/discussion.tex
\section{Discussion}\label{sec:discussion}
Our results show that the tested multimodal LLMs lack a basic perceptual understanding of simulated scenes.
Besides some understanding of simple shapes and colors, the models fail to answer basic questions about (relative) positions of static objects and movements of objects (Level 1). Because of that, it is not surprising that they also fail to judge the physical plausibility of simple object behaviors (Level 2).

Our model evaluation on Level 1 revealed that using a different prompting strategy (CoT) or changing the instruction (from yes-no questions to open-ended questions) can result in a substantial performance boost.
Specifically, the positive results using CoT prompting indicate that models are not able to extract necessary visual information following simple binary questions.
When ``guiding'' models with more unspecific questions first (initial prompt in CoT), they are sometimes able to extract the necessary information from their context to solve the same subsequent binary questions.
This highlights the high sensitivity to the nature of the prompts and the necessity for future models to improve upon these limitations. However, for Level 2, altering the prompting strategy did not impact the results. The models still failed to perform the task at this level, indicating that a lack of visual comprehension of more complex scenes is still at the heart of the problem.

Future work will encompass the analysis of additional multimodal LLMs.
Considering the recency of video-based multimodal LLMs (all the evaluated models were released last year), their capabilities may soon improve significantly.
Compared to text-based LLMs, which at this point contain hundreds of billions of parameters~\cite{naveed2023comprehensive}, video-based multimodal LLMs are at least one order of magnitude smaller~\cite{zhang2023videollama,su2023pandagpt,maaz2023videochatgpt}.
Supported by the observation that GPT-4 significantly outperforms Video-LLaMA and Video-LLaMA2 in tests with scene descriptions, an increase in model size alone might lead to the emergence of relevant language grounding and physical reasoning capabilities.
Image-based multimodal LLMs, such as GPT-4, have proven remarkably adept at answering complex and detailed questions about images.
Therefore, we aim to create an image-based version of GRASP and to compare image- and video-based models.
Throughout future developments in multimodal LLM capabilities, GRASP will prove instrumental in tracking the progression of these models, testing their grounding and physical comprehension capabilities against demanding data sets.

One potential reason for the poor performance of the models could be the discrepancy between the simulated videos in our benchmarks and the real-world training data of the models. In other words, our benchmark data is out-of-distribution (OOD) for the model. 
Generating a controlled dataset of real-world videos that test fundamental aspects of grounding and Intuitive Physics is difficult. Still, it would be interesting to conduct a comparison with selected examples of such stimuli in future work.
Either way, GRASP is useful as a challenging benchmark. 
Due to its OOD nature, it tests for scene understanding that generalizes to novel, and in our case abstracted, scenarios.
The limitations observed in current models' performance on GRASP's Level 1 stress the need for additional basic perceptual tests to allow for a more detailed analysis.
Furthermore, a future expansion to the benchmark could involve a subsequent level that requires models to address challenges within an \textit{interactive} simulated environment.
%Lastly, to disentangle failure attribution between the language and vision components of multimodal LLMs, we plan the inclusion of a comprehensive textual benchmark for Intuitive Physics.

%% file: sections/conclusion.tex
\section{Conclusion}\label{sec:conclusion}
GRASP introduces a robust grounding and Intuitive Physics benchmark tailored for multimodal LLMs.
By using simulated videos to model basic perceptual tasks and faithfully reproducing experiments from developmental psychology research within a simulation, GRASP serves as a comprehensive evaluation platform.
Results across both benchmark tiers demonstrate the challenging nature of GRASP. 
Notably, the results indicate a lack of perceptual understanding of simulated scenes by existing models, stressing the need for further development in this domain.
We plan to expand the benchmark in future work to facilitate research at the intersection of language and perception.

%% file: sections/acknowledgments.tex
\section*{Acknowledgments}
We would like to thank Dieuwke Hupkes for the idea that led to the creation of GRASP. We also thank Jülich Supercomputing Centre for the computing time and support to develop this project.

%% file: sections/contribution.tex
\section*{Contribution Statement}
Xenia Ohmer and Elia Bruni share senior authorship for this work.